\documentclass{Interspeech2024}
\interspeechfinaltrue
\PassOptionsToPackage{hyphens}{url}
\usepackage{arydshln}

\title{VisualSpeech: Enhancing Prosody Modeling in TTS Using Video}

\name{Shumin}{Que}
\name{Anton}{Ragni}
\address{
  Department of Computer Science, The University of Sheffield, United Kingdom}
\email{\{sque1, a.ragni\}@sheffield.ac.uk}


\begin{document}

\maketitle

\begin{abstract}
    Text-to-Speech (TTS) synthesis faces the inherent challenge of producing multiple speech outputs with varying prosody given a single text input. While previous research has addressed this by predicting prosodic information from both text and speech, additional contextual information, such as video, remains under-utilized despite being available in many applications. This paper investigates the potential of integrating visual context to enhance prosody prediction. We propose a novel model, VisualSpeech, which incorporates visual and textual information for improving prosody generation in TTS. Empirical results indicate that incorporating visual features improves prosodic modeling, enhancing the expressiveness of the synthesized speech. Audio samples are available at \href{https://ariameetgit.github.io/VISUALSPEECH-SAMPLES/}{\fontsize{7.4pt}{9pt}\tt https://ariameetgit.github.io/VISUALSPEECH-SAMPLES/}.
    
    % \url{https://ariameetgit.github.io/VISUALSPEECH-SAMPLES/}.
    %\url{https://anonymous.4open.science/w/VISUALSPEECH-SAMPLES-B222/}.

    % \href{https://ariameetgit.github.io/VISUALSPEECH-SAMPLES/}{https://ariameetgit.github.io/VISUALSPEECH-SAMPLES/}.
 
    % Manuscripts submitted to Interspeech 2024 must use this document as both an instruction set and as a template. Do not use a past paper as a template. Always start from a fresh copy, and read it all before replacing the content with your own. The main changes with respect to previous instructions are \red{highlighted in red}.
    
    % Before submitting, check that your manuscript conforms to this template. If it does not, it may be rejected. Do not be tempted to adjust the format! Instead, edit your content to fit the allowed space. The maximum number of manuscript pages is 5. The 5th page is reserved exclusively for \red{acknowledgements} and references, which may begin on an earlier page if there is space.
    
    % The abstract is limited to 1000 characters. The one in your manuscript and the one entered in the submission form must be identical. Avoid non-ASCII characters, symbols, maths, italics, etc as they may not display correctly in the abstract book. Do not use citations in the abstract: the abstract booklet will not include a bibliography.  Index terms appear immediately below the abstract. 
\end{abstract}

\section{Introduction}
With advancements in deep learning \cite{zen2014deep, zen2015unidirectional, tan2024naturalspeech}, text-to-speech (TTS) \cite{ren2019fastspeech,huang2022fastdiff} models have become capable of generating speech that closely mimics human speech. Despite the high-quality output, the prosody of the generated speech sometimes becomes inappropriate for the context. For example, most TTS approaches would fail to convey a different degree of happiness within the same sentence given a different context. This limitation arises because the same text input can produce multiple speech outputs with varying prosodic patterns. This is a well-known challenge in speech synthesis, known as the one-to-many mapping problem \cite{babianski2023granularity}.

Previous research has attempted to mitigate the one-to-many problem by incorporating additional inputs, including both textual and speech-related features. Commonly used text features encompass linguistic elements (e.g. part-of-speech, word boundaries, and word position) as well as syntactic structures and semantic meanings. In addition, some models incorporate an extra module to extract latent features, including pitch, energy, and duration, from reference speech \cite{ren2020fastspeech, kenter2019chive, valle2020mellotron}, capturing both global and local latent factors. Although these studies have enhanced prosodic modelling in TTS, there is still much work to be done.

In addition to text and speech, many common application scenarios, such as video game characters navigating dungeons or engaging in battles, feature visual information, which could play a crucial role in accurately predicting prosody. For instance, the speech patterns of these video game characters are unlikely to resemble those of read or acted speech due to the dynamic nature of visual contexts. There have so far been limited research on the usefulness of visual information in TTS. For instance, \cite{liu2023vit} successfully generated speech samples with accurate reverberation effects in specific environments by learning physical environment information from video data. However, no previous studies have explored the impact of visual information on prosody in TTS. 

Thus, this paper explores the possibility of improving prosody prediction in TTS with visual information (video). It makes three key contributions. First, it demonstrates visual cues carry valuable prosodic information. Second, this visual information complements existing textual features. Finally, it reveals that the integration of both textual and visual inputs leads to substantial improvements in prosody prediction in TTS models.

% \begin{enumerate}
%     \item We first explore whether visual features contain prosody information by predicting prosody-related features (including pitch and energy) conditioned on visual features extracted from two pre-trained visual models (i.e. the ResNet50 and Omnivore model). 

%     \item Furthermore, we propose a new VisualSpeech to generates speech conditioned on both visual and text inputs, which outperforms the baseline single-modal TTS model (i.e. Fastspech2) in terms of prosody. 

% \end{enumerate}

% \begin{figure}[t]
%   \centering
%   \includegraphics[width=\linewidth]{figure.pdf}
%   \caption{Schematic diagram of speech production.}
%   \label{fig:speech_production}
% \end{figure}

\begin{figure*}[t]
  \centering
  \includegraphics[width=\linewidth]{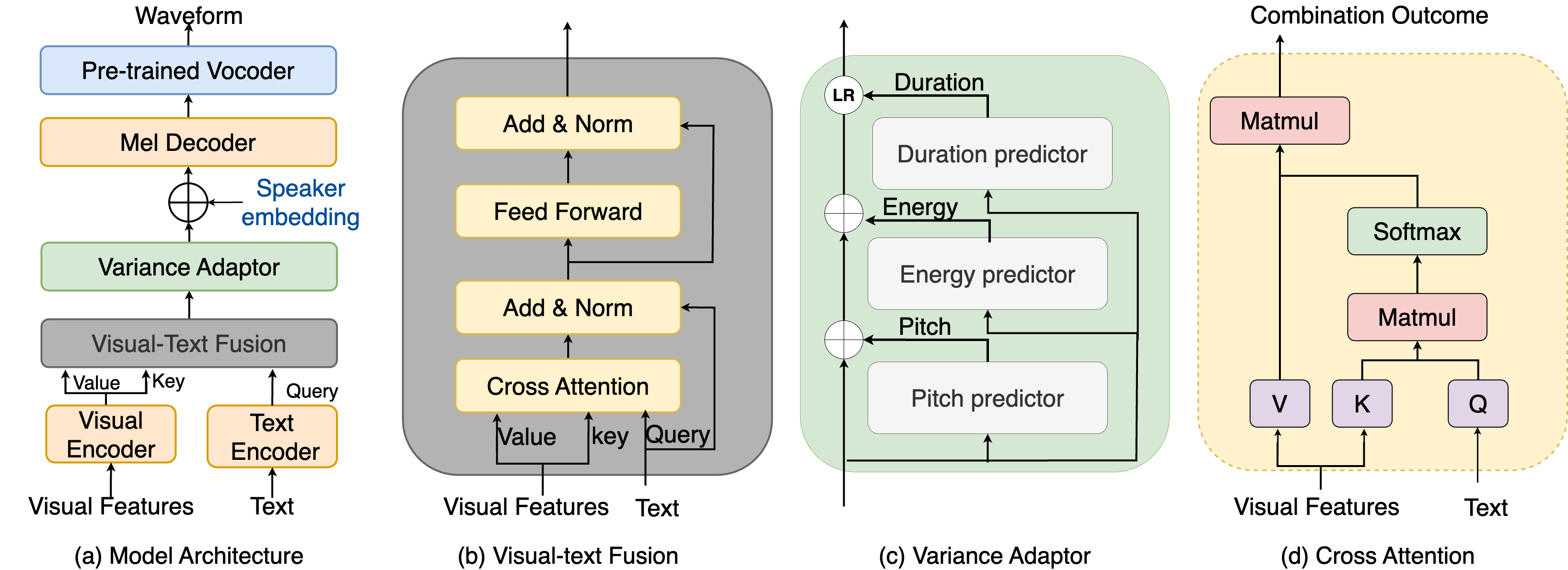}
  \caption{The proposed multi-modal TTS model: VisualSpeech.}
  \label{fig:visualspeech}
\end{figure*}

\section{The Proposed Method}
\subsection{Overview}
This paper first verifies that visual information is predictive of prosody. It then demonstrates that the prosodic information derived from visual features complements that extracted from text. Lastly, the visual features are integrated into a modern TTS model, as shown in Figure \ref{fig:visualspeech}. The architecture of the proposed model is based on FastSpeech2 \cite{ren2020fastspeech} due to its explicit prosody modeling and widespread adoption in the field. 

As shown in Figure \ref{fig:visualspeech}(a), the VisualSpeech model mainly consists of four standard FastSpeech2 modules (text encoder,  variance adaptor, Mel-spectrogram decoder, and pre-trained vocoder) and two new modules (visual encoder and visual-text fusion). Please refer to \cite{ren2020fastspeech} for details on the former modules. 

\subsection{The Visual Encoder}
The visual encoder shares a similar structure with the text encoder in FastSpeech2. Visual features, which refer to the characteristics extracted from video frames (such as facial expressions, body gestures, speaking environment, and other relevant visual cues), are represented as $\textbf{F} = \{\textbf{f}_1, \textbf{f}_2, \ldots, \textbf{f}_n\}$. These extracted features $\textbf{F}$ are then passed into the visual encoder to generate the visual hidden sequence $\textbf{V} = \{\textbf{v}_1,\textbf{v}_2, \ldots, \textbf{v}_n\}$.

\subsection{The Visual-Text Fusion}
\label{visual-Text}
Since the sequence lengths of the text encoder output $\textbf{T}$ and the visual encoder output $\textbf{V}$ are different, a simple addition or concatenation approach is infeasible in this case. Simple methods like average pooling can be adopted to compress $\textbf{V}$ into one global visual vector which can be added to the text encoder output $\textbf{T}$. However, local information might be lost if global visual vector is utilized due to the possibility of information over-compression. Thus, a more dedicated module ({\em i.e.} the visual-text fusion module) is proposed to better summarize the visual information and to shed light on how different parts of a video are correlated to the text. The proposed visual-text fusion module is illustrated in Figure \ref{fig:visualspeech}(b). It is designed to capture and leverage the complex interactions between textual and visual modalities. At its core, this module employs cross-attention that enables bidirectional information flow between the two modalities. This design choice is motivated by the observation that prosodic patterns often correlate with both linguistic content and visual context cues. 
The fusion architecture consists of two key components: (1) a cross-attention 
\begin{equation} 
    \delta(\mathbf{V}, \mathbf{T}) = \text{Softmax}\left(\frac{\mathbf{T}\mathbf{V}^T}{\sqrt{{d}_\mathbf{V}}}\right)\mathbf{V}
    \label{cross-attention}
\end{equation}
which is illustrated in Figure \ref{fig:visualspeech}(d), and (2) a feedforward network for feature refinement. In the cross-attention layer, text features are used as queries to attend to visual features, enabling the model to learn which visual cues are most relevant to specific textual elements and how these cues contribute to prosodic variation. Using the output of the text encoder $\mathbf{T}$ as the query can ensure that the output length of the visual-text fusion module is the same as the length of the phone embedding sequence, thus the output sequence of the module could be used to predict the pitch, energy, and duration at the phone level. $\mathbf{T}$ and $\mathbf{V}$ undergo a matrix-matrix product operation to obtain attention scores, which are divided by the square root of $d_{\mathbf{V}}$ ({\em i.e.} dimensionality of $\mathbf{V}$) and then passed through a Softmax function to get the normalized scores/weights. These normalized attention scores are used to obtain a weighted sum of $\mathbf{V}$.

% Figure \ref{fig:visualspeech}(d) and equation \ref{cross-attention} present the details of the cross-attention module. Specifically, the module takes the text encoder output $P$ as the query, and the visual encoder output $V$  as the key and value. Using the text encoder output $P$ as the query can ensure the output length of the visual-text fusion module is the same the phone embedding sequence length, thus the output sequence of the module could be used to predict the phone-level pitch, energy, and duration. $P$ and $V$ undergo a dot product operation to obtain the attention scores, which are divided by $d_v$ (i.e. the dimension of $V$) and then passed through a Softmax function to get the normalized scores/weights. These normalized attention scores are used to get a weighted sum of $V$. 

\section{Experiments}
\subsection{Experimental Setup}
To assess the impact of visual information on speech generation performance, a dataset containing both diverse prosodic and visual data is essential. However, research exploring the role of visual features in speech synthesis from a prosody perspective is still scarce, and existing datasets are limited in scope. While well-known speech synthesis datasets, such as LJSpeech \cite{ljspeech17} and LibriTTS \cite{zen2019libritts}, provide rich textual data, they lack corresponding visual features. On the other hand, datasets that do include visual information, like TED and Ego4D \cite{grauman2022ego4d}, suffer from limitations in prosodic or visual diversity. To address this gap, this paper utilizes the Condensed Movies Dataset (CMD) \cite{bain2020condensed}, an open-sourced, large-scale video dataset in English. CMD consists of short video clips extracted from over 3,605 films, spanning various genres, countries, and decades, making it a valuable resource for studying the integration of visual features in speech synthesis. These videos capture high-quality scenarios such as conversations at funerals and weddings, providing a rich array of prosodic and visual variations. 

The original duration of the video clips ranges from 10 to 300 seconds. Each clip was split into sentence segments by means of Whisper \cite{radford2023robust} to facilitate model training \cite{sanders2023cmd}. A subset of the resulting dataset was randomly selected for the experiments in this study, named CMD2, comprising approximately 33 hours of video and speech \cite{sanders2023cmd}. This subset includes 70,599 video clips ranging from 2 to 30 seconds along with corresponding speech and text transcriptions.% The resulting duration of video clips ranges from 2 to 30 seconds.
 
Compared to standard TTS datasets, CMD2 includes various types of background noise, music, and even some singing data. To prepare the CMD2 dataset for TTS training, this work applies vocal extraction techniques to separate music from speech, employs speech denoising and enhancement methods to reduce background noise, and implements filtering criteria to remove singing samples.

\begin{table}[t]
  \caption[The Objective Evaluations of the Original Speech and Preprocessed Speech]{The objective evaluations of the original speech (i.e. Raw), speech after vocal extraction (i.e. VE), and speech after vocal extraction and speech denoising and enhancement (i.e. VE + SDE). $\uparrow$ denotes the higher the better.}
  \label{table:ablation_result}
  \centering
  \begin{tabular}{l c c c} 
    \toprule
    \textbf{} & \textbf{STOI}$\uparrow$ & \textbf{PESQ} $\uparrow$ & \textbf{SI-SNR} $\uparrow$\\
    \midrule
    Raw   & 0.78 & 1.55 & 5.05 \\
    VE    & 0.84 & 1.85 & 9.06 \\
    VE+SDE &0.95 & 2.90 & 16.49\\
    \bottomrule
  \end{tabular}  
\end{table}

This work utilizes MVSEP \cite{fabbro2023sound} for vocal separation and Resemble-Enhance \cite{resemblyzer} for speech denoising and enhancement. Objective evaluations were conducted to assess the impact of vocal extraction, speech denoising, and enhancement on the speech quality. Following the methodology in \cite{vyas2023audiobox}, the evaluation metrics include Short-Time Objective Intelligibility~(STOI) \cite{taal2010short}, Wideband Perceptual Evaluation of Speech Quality~(PESQ) \cite{rix2001perceptual}, and Scale-Invariant Signal-to-Distortion Ratio ~(SI-SDR) \cite{le2019sdr}. As reference (clean) speech is unavailable, reference-free scores for these metrics were estimated using the method in \cite{kumar2023torchaudio}. A random selection of 1,000 samples from the dataset was used to compute these metrics. The results, presented in Table \ref{table:ablation_result}, clearly demonstrate that, compared to the raw dataset, vocal extraction yields relative improvements of 7.7\%, 19.4\%, and 79.4\%, while speech denoising and enhancement provide even more substantial improvements, highlighting the effectiveness of both techniques.

The following filtering criteria are applied to remove singing clips from the dataset. First, speech samples with pitch values exceeding a predefined threshold (i.e., 500 Hz) are discarded. Second, we compute the mean and standard deviation of pitch, energy, and duration for each phoneme. Any speech sample with pitch or duration values that fall outside 2.5 times the standard deviation of these statistics is considered an outlier and is removed. After applying these filters, the resulting dataset(44,665 training, 2,000 validation, and 200 test samples) is prepared for training the TTS model.

Following the common practice in \cite{ren2019fastspeech}, we resampled all speech into 22.05k sample rate, and converted it into 80-dimensional Mel-spectrograms. We used the Montreal Forced Aligner (MFA) to perform forced alignment and extract phoneme duration.

% Statistical methods are used to identify and remove outliers. Additionally, specific steps are taken to eliminate clips associated with singing.

\subsection{Preliminary Study I: Prosody Clues in Visual Features}

To verify the hypothesis of visual features in prosody modeling, a preliminary experiment is conducted using a FeedForward Neural Network (FFNN) to predict prosodic features (pitch and energy), solely from visual features. 

Given a sequence of visual features $\mathbf{V} = \{\mathbf{v}_1, \mathbf{v}_2,\ldots,\mathbf{v}_n\}$ and a sequence of prosodic features $\mathbf{P} = \{\mathbf{p}_1, \mathbf{p}_2,\ldots,\mathbf{p}_t\}$, where \(t\gg n\), we average every \(t/n\) prosody features to obtain a target prosody feature corresponding to each visual feature. As a result, the final prosody feature sequence $\mathbf{P}_{\mathbf{V}} = \{\mathbf{p}_{\mathbf{v}_1}, \mathbf{p}_{\mathbf{v}_2}, \ldots, \mathbf{p}_{\mathbf{v}_n}\}$ aligns with the visual feature sequence. In this setup, the visual feature sequence $\mathbf{V}$ is used as the input to the model, and the prosody sequence $\mathbf{P}_{\mathbf{V}}$ serves as the target. %As the duration of each visual feature is fixed, there is no need for duration prediction. 

In this experiment, two types of visual features were investigated: one extracted from the Omnivore model \cite{girdhar2022omnivore}\footnote{\url{https://github.com/facebookresearch/omnivore}} and the other from the ResNet50 model \cite{he2016deep}.\footnote{\url{https://github.com/v-iashin/video_features}\\$^\dagger$ The results in the accepted paper contain a typographical error.} The FFNN models consist of two hidden layers, each with 256 dimensions and a dropout rate of 0.5. Each model was trained for 400,000 steps with a batch size of 32. The loss function used is mean squared error (MSE). The Adam optimizer \cite{kingma2014adam} was employed for parameter updates, with hyperparameters \(\beta_1 = 0.9\), \(\beta_2 = 0.98\), and a learning rate of \(1 \times 10^{-5}\).

% \begin{table}[h]
% \centering
% \resizebox{\textwidth}{!}{
% \begin{tabular}{lccc}
% \toprule 
% \textbf{} &  \textbf{Valid(P/E)} \downarrow & \textbf{Test(P/E)} \downarrow \\
% \midrule
% MSE(ref, FNN-ResNet50)  &  1.041/1.192 & 0.929/1.199  \\\hline
% MSE(ref, FNN-Omnivore)  &  0.944/1.099 & 0.844/1.112 \\
% \bottomrule
% \end{tabular}

% \begin{table}[t]
%   \caption[MSE Loss of FNN-ResNet50 and FNN-Omnivore Models]{MSE Loss of FNN-ResNet50 and FNN-Omnivore Models on Validation and Test Datasets}
%   \label{tab:FNN_Predictive}
%   \centering
%   \begin{tabular}{l c c} % 定义了 4 列
%   %\begin{tabular}{ r@{}l  r }
%     \toprule
%     \textbf{Features} &  \textbf{Valid(P/E)} $\downarrow$ & \textbf{Test(P/E)} $\downarrow$ \\
%     %\multicolumn{3}{c}{1} & {\textbf{Valid(P/E)}} & {\textbf{Text(P/E)}} \\
%     \midrule
%     MSE(ref, FNN-ResNet50)           & 1.041/1.192 &  0.929/1.199 \\
%     MSE(ref, FNN-Omnivore)           & 0.944/1.099 & 0.844/1.112 \\   
%     \bottomrule
%   \end{tabular} 
% \end{table}

\begin{table}[t]
  \caption[MSE Loss of FNN-ResNet50 and FNN-Omnivore Models]{MSE Loss of FNN-ResNet50 and FNN-Omnivore Models on Test Datasets.}
  \label{tab:FNN_Predictive}
  \centering
  \begin{tabular}{l c c} % 定义了 4 列
  %\begin{tabular}{ r@{}l  r }
    \toprule
    \textbf{Models} &  \textbf{Pitch} $\downarrow$ & \textbf{Energy} $\downarrow$ \\
    %\multicolumn{3}{c}{1} & {\textbf{Valid(P/E)}} & {\textbf{Text(P/E)}} \\
    \midrule
    Mean-predictor  & 2.16$^\dagger$ & 3.31$^\dagger$ \\
    \hline
    FNN-ResNet50           & 0.93 &  1.20 \\
    FNN-Omnivore           & 0.84 &  1.11 \\   
    \bottomrule
  \end{tabular} 
\end{table}

% \begin{table}[t]
%   \caption[MSE Loss of FNN-ResNet50 and FNN-Omnivore Models]{MSE Loss of FNN-ResNet50 and FNN-Omnivore Models on Test Datasets.}
%   \label{tab:FNN_Predictive}
%   \centering
%   \begin{tabular}{l c c} % 定义了 4 列
%   %\begin{tabular}{ r@{}l  r }
%     \toprule
%     \textbf{Type} &  \textbf{Pitch}  & \textbf{Energy} \\
%     %\multicolumn{3}{c}{1} & {\textbf{Valid(P/E)}} & {\textbf{Text(P/E)}} \\
%     % \midrule
%     % Mean-predictor  & 23.72 & 28.97 \\
%     \hline
%     Mean           & 123.37 &52.78\\
%     Std          & 131.96 &  39.94 \\
%     Example(ori)          & 101.83 &  68.82 \\
%     Example(after)          & -0.1659 &  0.3964 \\
%     \bottomrule
%   \end{tabular} 
% \end{table}

The results are presented in Table \ref{tab:FNN_Predictive}. For reference, we also include the results from a mean predictor, where the predicted value is simply the mean of the ground truth sequence, used to compute the MSE loss. The MSE loss for the model using visual features is significantly smaller than that of the mean predictor, clearly demonstrating that visual features are predictive of prosody. Moreover, the MSE loss obtained using visual features from the Omnivore model is slightly smaller than that from ResNet50, indicating that Omnivore extracts more relevant information for prosody prediction than ResNet50.

\subsection{Preliminary Study II: Visual Features as a Complement to Textual Information}

Having confirmed visual information can predict prosody, this section investigates whether visual features complement textual features. To this end, we extend the pitch, energy, and duration (PED) predictor in FastSpeech2 \cite{ren2020fastspeech} to incorporate both textual and visual information. We then compare performance of the model using both types of features with that of the model using only textual features. This comparison allows us to assess the added value of visual features in enhancing prosody prediction.

The text-only PED predictor, following the methodology outlined in \cite{ren2020fastspeech}, relies solely on textual input for prosody prediction. In contrast, the visual-textual PED predictor integrates both textual and visual features to predict prosody. The prosodic features—pitch, energy, and duration (P/E/D)—are extracted from the ground-truth (original) audio and serve as the target for both predictors. The visual encoder shares the same architecture as the text encoder. The cross-attention module in the visual-text fusion model has a hidden size of 256, with two attention heads and a dropout rate of 0.2. The training setup follows that of \cite{ren2020fastspeech}, with the exception of a batch size of 32 and a total of 400,000 training steps.  The Adam optimizer \cite{kingma2014adam} is used for parameter updates, with hyperparameters \(\beta_1 = 0.9\), \(\beta_2 = 0.98\), and a learning rate scheduler, as specified in \cite{ren2020fastspeech}.

\begin{table}[t]
  \caption[MSE Loss of Three Models (PED)]{MSE Loss of three models (Text-based, Text + Omnivore Visual Features, and Text + ResNet50 Visual Features) on test datasets.}
  \label{table:prediction_results_ped}
  \centering
  \begin{tabular}{l c c c} 
    \toprule 
    \textbf{Models} &  \textbf{Pitch} $\downarrow$ & \textbf{Energy} $\downarrow$  & \textbf{Duration} $\downarrow$ \\
    \midrule
     Text              & 0.43 & 0.50 & 0.35\\
     Text+VF-Omnivore  & 0.39 & 0.59 & 0.34\\
     Text+VF-ResNet50  & 0.40 & 0.58 & 0.35 \\
    \bottomrule
  \end{tabular} 
\end{table}

The results presented in Table \ref{table:prediction_results_ped} show that the inclusion of visual features improves pitch prediction slightly, with the Text + VF-Omnivore model achieving the lowest MSE loss (0.39). In terms of duration prediction, the Text + VF-Omnivore model achieves a relative 3\% improvement compared with the text-only model, and the Text + VF-ResNet50 model gains no improvement. However, for energy prediction, the addition of visual features has a negative impact on performance.

\subsection{TTS Experiments}
Previous results demonstrate that the combination of visual and textual information outperforms text-only prosody prediction, including pitch and duration. To verify whether these improvement in prosody feature prediction will lead to better speech generation, visual features are integrated into the FastSpeech2 \cite{ren2020fastspeech} model,\footnote{ Experiments are implemented based on the open-sourced repository:  \href{https://github.com/ming024/FastSpeech2}{\fontsize{7.4pt}{9pt}\tt https://github.com/ming024/FastSpeech2}} incorporating a visual encoder and visual-text fusion, to synthesize speech. Two VisualSpeech models were trained, one with Omnivore and another with ResNet50 visual features. Each model is trained for 400,000 steps with a batch size of 32. The Adam optimizer is used for parameter updates \cite{kingma2014adam}, with hyper-parameters \(\beta_1 = 0.9\), \(\beta_2 = 0.98\), and the same learning rate scheduler as \cite{ren2020fastspeech}. 

% The loss function used in this experiment is the mean square error (MSE), as follows

% \begin{equation}
%     L = \frac{1}{N} \sum_{i=1}^{N} (P_i - \hat{P}_i)^2 +  (E_i - \hat{E}_i)^2 + (D_i - \hat{D}_i)^2 + (M_i - \hat{M}_i)^2 
% \end{equation}
% where N is the total number of training data, $P_i$,  $E_i$,  $D_i$, and $M_i$ refer to the ground truth pitch, energy, duration, mel-spectrogram, respectively, $\hat{P}_i$, $\hat{E}_i$, $\hat{D}_i$, and $\hat{M}_i$ denote the corresponding predicted values.

\begin{table}[t]
  \caption[MSE Loss of Three Models (TTS)]{MSE loss of three Models (FastSpeech2, VisualSpeech with Omnivore Features, and VisualSpeech with ResNet50 Features) on Test Datasets}
  \label{table:prediction_results_tts}
  \centering
  \begin{tabular}{l c c c} 
    \toprule
    \textbf{Models} &  \textbf{Pitch} $\downarrow$ & \textbf{Energy} $\downarrow$  & \textbf{Duration} $\downarrow$ \\
    \midrule
     FastSpeech2                & 0.27 & 0.39 &0.41\\
     VisualSpeech (Omnivore)    &   0.18 &0.37 &0.21\\
     VisualSpeech (ResNet50)    &  0.18 & 0.34 & 0.24\\
    \bottomrule
  \end{tabular} 
\end{table}

% \begin{table}[t]
%   \caption[The Mel-cepstral Distance (MCD) (in dB) and Log F0 RMSE Loss of Three Models]{The Mel-cepstral Distance (MCD) (in dB) and Log F0 RMSE Loss of Three Models}
%   \label{table:prediction_results_mcd_log}
%   \centering
%   \begin{tabular}{l c c c} 
%     \toprule
%     \textbf{Models} & \textbf{MCD}$\downarrow$ & \textbf{Log F0}$\downarrow$ \\
%     \midrule
%     FastSpeech2            & 7.64 & 0.46 \\
%     VisualSpeech (Omnivore)  & 7.38 & 0.44\\
%     VisualSpeech (ResNet50) & 7.34 & 0.45 \\
%     \midrule 
    
%     FastSpeech2 (+ GT PED)            & 5.04 & 0.32 \\
%     VisualSpeech (Omnivore) (+ GT PED)    & 4.83 & 0.30\\
%     VisualSpeech (ResNet50) (+ GT PED)   & 4.82 & 0.31 \\
%     \bottomrule
%   \end{tabular}  
% \end{table}

 % \footnote{Relative improvement refers to the percentage improvement compared to the baseline model. It is calculated as: $\frac{\text{Baseline Metric} - \text{Proposed Model Metric}}{\text{Baseline Metric}} \times 100\%$. For example, if the baseline pitch MSE is 0.27 and the proposed model's pitch MSE is 0.18, the relative improvement is $\frac{0.27 - 0.18}{0.27} \approx 0.33$, or 33\%.}

Table~\ref{table:prediction_results_tts} shows that the proposed models significantly outperform the baseline FastSpeech2 model on all three metrics. The Omnivore-based model achieves about 33\%, 5\%, and 49\% relative improvement in pitch, energy, and duration prediction, respectively. We find that the improvement in the energy prediction is relatively less significant, which may be attributed to energy being less predictable. FastPitch \cite{lancucki2021fastpitch} also found removing energy prediction while maintaining pitch and duration prediction could generate high-quality speech.

Table \ref{table:prediction_results_mcd_log}  demonstrates that the proposed model significantly outperforms the baseline model in terms of Mel-Cepstral Distortion (MCD) and UTMOS \footnote{UTMOS \href{https://github.com/sarulab-speech/UTMOS22}{\fontsize{7.4pt}{9pt}\tt https://github.com/sarulab-speech/UTMOS22} results for these models (i.e. 2.91, 3.06 and 3.13, respectively) suggest that these improvements have led to an enhanced perceptual quality.} and offers slight improvement in Log F0. Similar trends are observed when using ground truth (GT) PED, suggesting that visual features also influence mel-spectrogram prediction beyond PED. These findings indicate that the current approach extracts some useful prosodic information from video, but there is still significant room for improvement. The remaining gap in MCD suggests that further refinement in the modeling of visual features or a more effective utilization of the available data could enhance their contribution to prosody modeling, extending beyond PED prediction alone.

\begin{table}[t]
  \caption[The Mel-cepstral Distance (MCD) (in DB) and Log F0 RMSE Loss of Three Models]{The Mel-cepstral Distance (MCD) (in DB) and Log F0 RMSE Loss of Three Models}
  \label{table:prediction_results_mcd_log}
  \centering
  \begin{tabular}{l c c c} 
    \toprule
    \textbf{Models} & \textbf{MCD}$\downarrow$ & \textbf{Log F0}$\downarrow$ \\
    \midrule
    FastSpeech2            & 7.64 & 0.46 \\
    VisualSpeech (Omnivore)  & 7.38 & 0.44\\
    VisualSpeech (ResNet50) & 7.34 & 0.45 \\
    \midrule 
    
    FastSpeech2 (+ GT PED)            & 5.04 & 0.32 \\
    VisualSpeech (Omnivore) (+ GT PED)    & 4.83 & 0.30\\
    VisualSpeech (ResNet50) (+ GT PED)   & 4.82 & 0.31 \\
    \bottomrule
  \end{tabular}  
\end{table}

\subsubsection{Case Study: Prosody Visualization}
To better understand the differences between models trained in the TTS experiments and evaluate their performance on test data, we used these three well-trained models to generate speech given the same text input and visualize them. 

\begin{figure}[t]
    \centering
    \includegraphics[width=\linewidth]{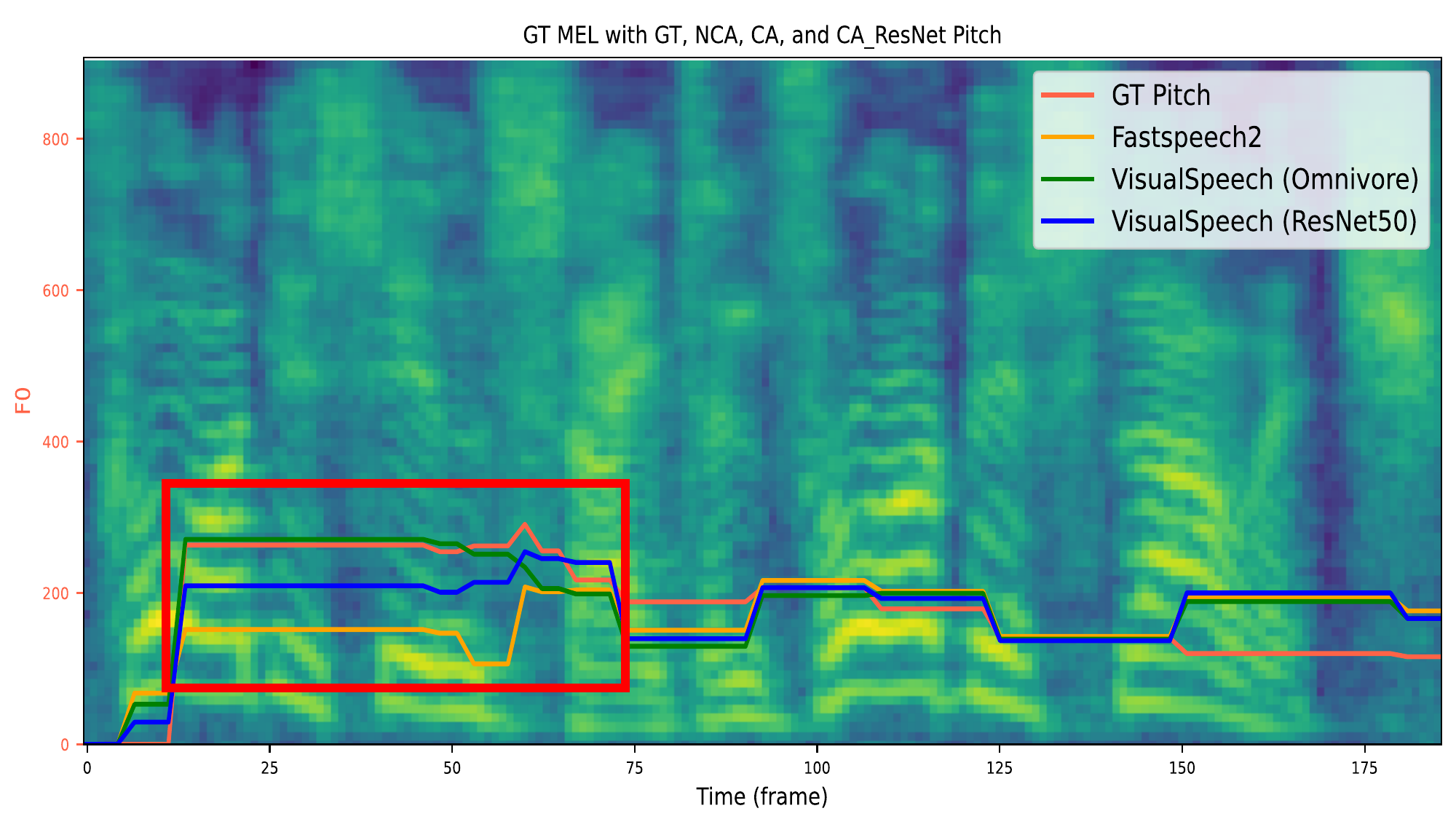}
    \caption{The Pitch Contour}
    \label{fig:pitch_contout}
\end{figure}

\begin{figure}[t]
    \centering
    \includegraphics[width=1.05\linewidth]{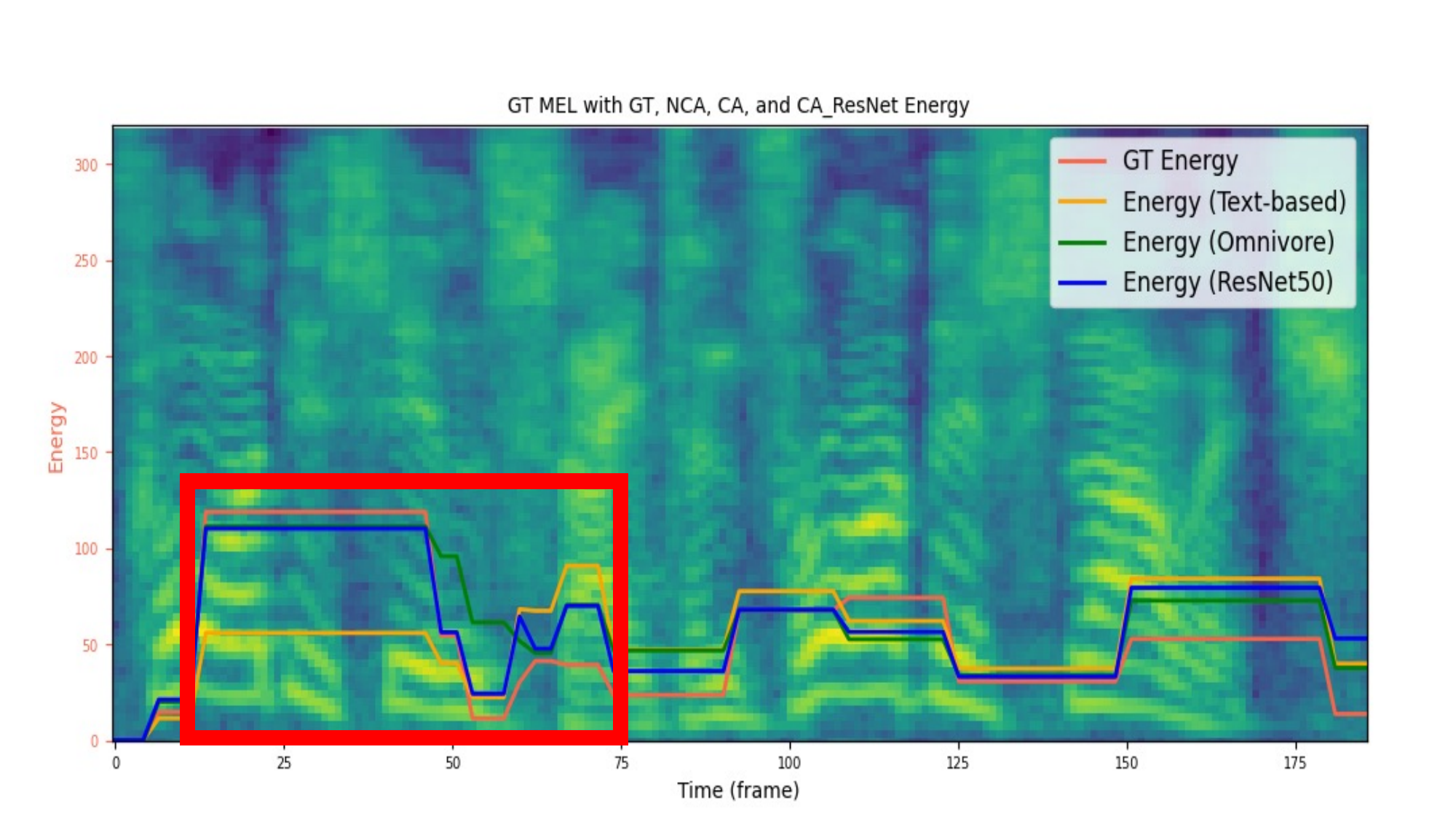}
    \caption{The Energy Contour}
    \label{fig:energy_contout}
\end{figure}

% \begin{figure}[t]
%     \centering
%     \includegraphics[width=1.2\linewidth]{duration.pdf}
%     %\includegraphics[width=1.5\linewidth]{duration.pdf}
%     \caption[Visualization of Mel-spectrogram Generated By Various Model]{Visualization of Mel-spectrogram generated by various model.}
%     \label{fig:duration_mel}
% \end{figure}

% In Figure \ref{fig:duration_mel}, we visualize the mel-spectrograms generated by different models. The closer a mel-spectrogram's length is to the ground truth (GT), the better the model performs in duration prediction. The figure clearly shows that the mel-spectrograms generated by VisualSpeech (Omnivore) and VisualSpeech (ResNet50) are much closer in length to the GT compared to the one generated by FastSpeech2. This indicates that VisualSpeech outperforms FastSpeech2 in duration prediction. 

In Figure \ref{fig:pitch_contout}, we visualize the predicted pitch contours from three models and compare them with the ground truth. As shown, the pitch contours generated by the two models incorporating additive visual features are closer to those produced by the text-only model. A similar trend is observed in energy prediction, as illustrated in Figure \ref{fig:energy_contout}.

These findings align with other studies that have demonstrated the use of visual features to improve speech across various dimensions \cite{xie2024sonicvisionlm, liu2023vit, mira2022end, choi2023intelligible}, further confirming the effectiveness of visual features in speech synthesis. Additionally, the improvement in prosodic prediction through visual features is consistent with the discussion in \cite{hodari2021camp}, which suggested that contextual information enhances prosody in generated speech. As previously mentioned, visual features provide additional context, such as the speaking environment, which plays a critical role in more accurately predicting prosody.

% \subsection{Attention Weight Visualization}
% To better understand the relationship between visual features and text, we visualize the attention weights between visual features and phoneme sequences in the cross-attention module. The comparison between two different models, Omnivore and ResNet50, is presented in Figure \ref{fig:res_weight} and  \ref{fig:om_weight}. In these visualizations, lighter pixels indicate higher attention weights, suggesting a stronger correlation.

% Both heatmaps show that the cross-attention module tends to focus more on the initial and final parts of the visual hidden states, particularly in the case of the Omnivore model. This pattern suggests that these visual features may capture more `global' aspects of speech, rather than fine-grained details. This may be attributed to the fact that these models are trained on classification tasks that emphasize global features over subtle differences across individual video frames.

% \begin{figure}[t]
%     \centering
%     \includegraphics[width=1\linewidth]{tt0020629_0_368Nrtkmrls_22_attention.pdf}
%     %\includegraphics[width=1.5\linewidth]{duration.pdf}
%     \caption{The Omnivore model}
%     \label{fig:om_weight}
% \end{figure}

% \begin{figure}[t]
%     \centering
%     \includegraphics[width=1\linewidth]{tt0020629_0_368Nrtkmrls_22_attention_resnet.pdf}
%     %\includegraphics[width=1.5\linewidth]{duration.pdf}
%     \caption{The ResNet50 model}
%     \label{fig:res_weight}
% \end{figure}

\section{Conclusion}
In this work, we introduce VisualSpeech, the first visual text-to-speech synthesis model that incorporates visual features from corresponding videos to complement text features, significantly enhancing the prosody of the generated speech. Using two distinct video feature extractors, we demonstrate that these visual features encapsulate prosodic information. By integrating these features with text, our results show that visual information complements textual information, leading to improved prosody prediction. This study lays a foundation for future research on leveraging visual information to improve prosody performance in TTS. Additionally, the CMD2 dataset created in this work provides a valuable resource for future studies.

% Looking ahead, effectively incorporating visual information to improve prosody in speech synthesis remains a critical objective. 

While this study has shown promising results, there are some limitations. Despite employing speech denoising and enhancement, the quality of the CMD dataset still impacts the speech generation quality. Therefore, a high-quality dataset is essential for future research.

%The use of a pre-trained vocoder also limits the generation quality; training or fine-tuning a vocoder specifically on the target dataset could further enhance the results.

% This will pave the way for further advancements in prosody modeling and synthesis.

% As indicated in the experimental setup, we utilized an adjusted Omnivore model and ResNet50 to extract visual features, future work could explore the original Omnivore model or newer extractors for improved performance. 

% \section{Acknowledgements}
% Acknowledgement should only be included in the camera-ready version, not in the version submitted for review.
% The 5th page is reserved exclusively for \red{acknowledgements} and  references. No other content must appear on the 5th page. Appendices, if any, must be within the first 4 pages. The acknowledgments and references may start on an earlier page, if there is space.

% \ifinterspeechfinal
%      The Interspeech 2024 organisers
% \else
%      The authors
% \fi
% would like to thank ISCA and the organising committees of past Interspeech conferences for their help and for kindly providing the previous version of this template.

\bibliographystyle{IEEEtran}
\bibliography{mybib}

\end{document}